\pdfoutput=1

\documentclass[11pt]{article}

\usepackage[final]{acl}

\usepackage[utf8]{inputenc} 
\usepackage[T1]{fontenc}
\usepackage{graphicx}
\usepackage{amsmath,amsfonts}
\usepackage{microtype}
\usepackage{inconsolata}
\usepackage{nicefrac}
\usepackage{booktabs}
\usepackage{hyperref}
\usepackage{times}
\usepackage{latexsym}
\usepackage{url}
\usepackage{graphicx}
\usepackage{amsmath}
\usepackage{amssymb}
\usepackage{subcaption}

\usepackage[T1]{fontenc}
\usepackage[utf8]{inputenc}
\usepackage{microtype}

\title{ConciseRL: Conciseness-Guided Reinforcement Learning for Efficient Reasoning Models}

\author{
Razvan-Gabriel Dumitru \\
University of Arizona \\
ServiceNow AI \\
\texttt{razvandumm@gmail.com}
\And
Darius Peteleaza \\
MultiversX \\
Lucian Blaga University of Sibiu \\
\texttt{peteleaza.darius@gmail.com}
\AND
Vikas Yadav \\
ServiceNow AI
\And
Liangming Pan \\
University of Arizona
}

\begin{document}
\maketitle

\begin{abstract}
Large language models excel at complex tasks by breaking down problems into structured reasoning steps. However, reasoning traces often extend beyond reaching a correct answer, causing wasted computation, reduced readability, and hallucinations. To address this, we introduce a novel hyperparameter-free conciseness score used as a reward signal within a reinforcement learning framework to guide models toward generating correct and concise reasoning traces. This score is evaluated by a large language model acting as a judge, enabling dynamic, context-aware feedback beyond simple token length. Our method achieves state-of-the-art efficiency–accuracy trade-offs on the MATH dataset, reducing token usage by up to 31$\times$ on simple problems while improving accuracy by 7\%, and on the hardest problems, it outperforms full reasoning by +7.5\% accuracy with up to 3.6$\times$ fewer tokens. On TheoremQA, our method improves accuracy by +2.2\% using 12.5$\times$ fewer tokens. We also conduct ablation studies on the judge model, reward composition, and problem difficulty, showing that our method dynamically adapts reasoning length based on problem difficulty and benefits significantly from stronger judges. The code, model weights, and datasets are open-sourced at https://github.com/RazvanDu/ConciseRL.
\end{abstract}

\section{Introduction}

Large language models (LLMs) have recently made significant progress in solving complex multi-step tasks, such as mathematics, code generation, and symbolic reasoning. Reasoning models such as o1 \cite{openaio1}, DeepSeek-R1 \cite{deepseekai2025deepseekr1incentivizingreasoningcapability}, and S1 \cite{muennighoff2025s1simpletesttimescaling} are trained to reason explicitly through intermediate steps, often using reinforcement learning (RL) to optimize for correctness and structural fidelity. While this explicit reasoning improves accuracy and interpretability, it also introduces a major challenge: reasoning traces tend to be excessively long \cite{chen2025think23overthinkingo1like, kimiteam2025kimik15scalingreinforcement}. These models frequently continue reasoning well past the point where the correct answer is reached, resulting in wasted computation, degraded readability, and sometimes even contradictions or hallucinated steps. Addressing this problem is essential for reducing inference costs and improving the usability of LLM-generated reasoning in real-world applications.

\begin{figure*}[ht]
    \includegraphics[width=\linewidth]{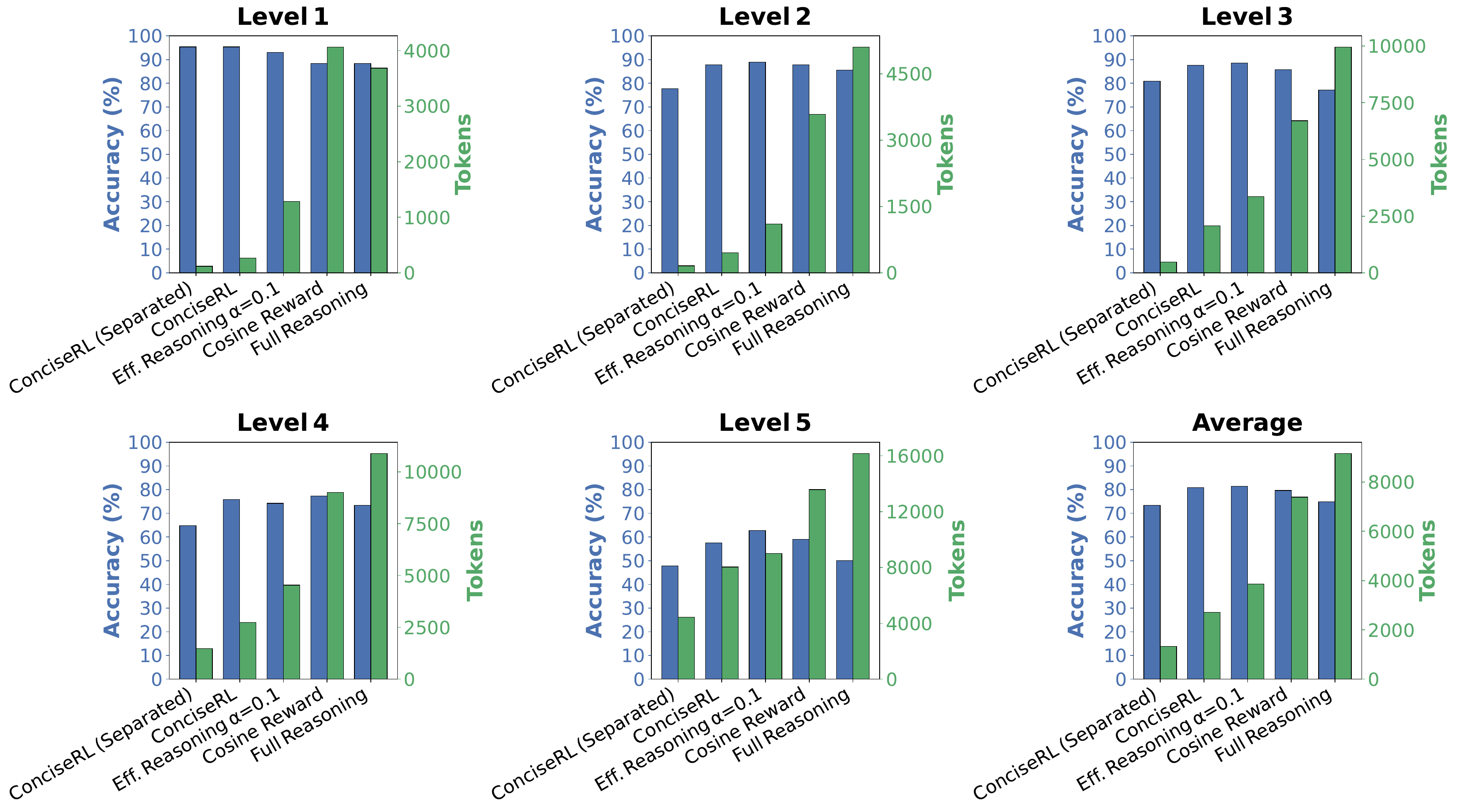}
    \caption{MATH500 histogram by difficulty level. We report both \textcolor{blue}{accuracy} (blue, left axis) and average \textcolor{green}{token length} (green, right axis) for each method. All methods are based on DeepSeek-R1-Distill-Qwen-1.5B. For our method ("ConciseRL" and "ConciseRL (Separated)"), we use GPT-4.1 mini as the judge. The exact values shown in the histogram are reported in Table~\ref{tab:math_difficulty_levels}.}
    \label{fig:histogram}
\end{figure*}

In this work, we propose a novel approach to reduce excessive reasoning in LLMs by teaching them to generate answers that are both correct and concise. Our method introduces a semantically informed \textbf{conciseness score} as a reward function for training reasoning models and an LLM as a judge to evaluate the conciseness of the reasoning trace. One key insight is that while accuracy can often be judged deterministically (e.g., via matching answers or symbolic execution), conciseness is inherently subjective and better suited for LLM-based evaluation. 

\begin{figure*}[ht]
    \includegraphics[width=\linewidth]{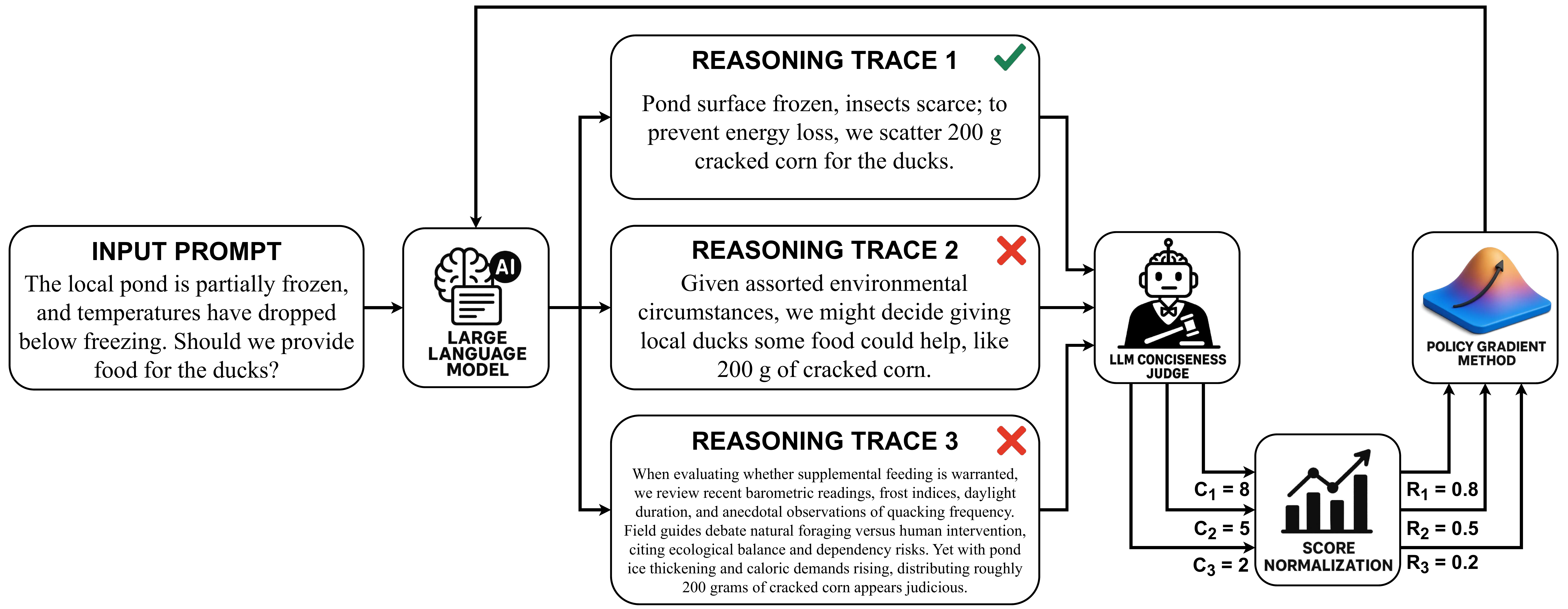}
    \caption{Given an input prompt, an LLM generates multiple reasoning traces that are evaluated by an LLM-based judge who scores each trace based on \textbf{conciseness}. Trace 1 is concise and receives the highest reward, Trace 2 has an equal length (24 tokens) but lower conciseness, while Trace 3 is the longest (71 tokens) and least concise. These rewards then guide a policy gradient update.}
    \label{fig:workflow}
\end{figure*}

Previous work has explored LLM-as-a-judge setups for evaluating factual accuracy, style, and helpfulness \cite{li2025generationjudgmentopportunitieschallenges, NEURIPS2023_91f18a12}, but existing methods for controlling reasoning length rely on token count or static heuristics, which fail to capture semantic efficiency. Our method goes beyond simple token count: concise traces are often short, but short traces are not always concise, therefore, the reward targets conciseness directly, and a shorter length appears only as a side effect. As illustrated in Figure~\ref{fig:workflow}, all three examples use the same number of tokens, yet the first is the most concise, receiving the highest reward. The second and third are progressively less concise, resulting in lower rewards. Unlike static length penalties, our LLM-based conciseness score is dynamic and context-aware, enabling better generalization across problem types and difficulty levels while improving explainability. Additionally, our method can be combined with correctness-based rewards, offering a flexible trade-off between conciseness and robustness.

\paragraph{Our main contributions are as follows:}
\begin{itemize}
    \item We introduce a novel \textbf{conciseness score} and the first method that leverages an LLM to evaluate and provide a reward score based on the \textbf{conciseness} of reasoning traces. Unlike prior token-level heuristics that require careful hyperparameter tuning, our method captures the semantic compactness of reasoning in a context-sensitive and dynamic way, \textbf{with no additional hyperparameters.}
    \item Our method enables adaptive reasoning by adjusting response length based on problem difficulty. As shown in Figure~\ref{fig:histogram} and Table~\ref{tab:math_difficulty_levels}, it allocates more tokens to harder MATH500 problems while using \textbf{31$\times$ fewer tokens} on easier ones and improving accuracy by \textbf{7\%}. On the hardest questions, it outperforms full reasoning by \textbf{7.5\% accuracy} with \textbf{3.6$\times$ fewer tokens}. On TheoremQA, our method achieves \textbf{2.2\% higher accuracy} using \textbf{12.5$\times$ fewer tokens} than the full reasoning model. This highlights a level of length-efficiency control not achieved by any static heuristic or prior method.
    \item We perform extensive experiments and ablations across multiple benchmarks, including conciseness-only and accuracy-combined rewards, conditioning on problem difficulty, and analyzing how different judge models affect training dynamics, reward signals, and reasoning quality.
\end{itemize}

\section{Related Work}

Chain-of-Thought (CoT) \cite{NEURIPS2022_9d560961} enhances LLM reasoning by prompting explicit intermediate steps, improving both performance and interpretability in complex tasks \cite{snell2024scalingllmtesttimecompute, NEURIPS2022_b1efde53}. Variants such as Tree-of-Thought \cite{NEURIPS2023_271db992} and Graph-of-Thought  \cite{Besta_Blach_Kubicek_Gerstenberger_Podstawski_Gianinazzi_Gajda_Lehmann_Niewiadomski_Nyczyk_Hoefler_2024} extend the CoT paradigm through structured search and iterative refinement. Self-Consistency prompting \cite{wang2023selfconsistencyimproveschainthought} improves robustness by sampling and aggregating multiple reasoning paths but at the cost of increased computational overhead from longer traces.

Recent advances in reasoning models such as o1 \cite{openaio1}, o3 \cite{openaio3o4mini}, o4-mini \cite{openaio3o4mini}, DeepSeek-R1 \cite{deepseekai2025deepseekr1incentivizingreasoningcapability}, S1 \cite{muennighoff2025s1simpletesttimescaling}, and QwQ-32B \cite{qwenqwq32b} show strong reasoning through internal capabilities, without inference-time prompting. Some methods adopt tree-based search strategies \cite{yao2024mulberryempoweringmllmo1like}, while others, like DeepSeek-R1 \cite{deepseekai2025deepseekr1incentivizingreasoningcapability, shao2024deepseekmathpushinglimitsmathematical}, combine supervised fine-tuning with RL. Despite their performance, these methods are prone to generating verbose reasoning traces, a phenomenon known as the overthinking problem \cite{chen2025think23overthinkingo1like, kimiteam2025kimik15scalingreinforcement}.

Overthinking occurs when a model continues generating redundant reasoning steps after reaching the correct answer. This behavior wastes inference compute and risks introducing inconsistencies and logic errors \cite{chen2025think23overthinkingo1like}. This issue is amplified by training objectives that reward long reasoning sequences \cite{deepseekai2025deepseekr1incentivizingreasoningcapability}, creating a misalignment between performance and efficiency. 

To address these issues, several approaches have been proposed for CoT compression and length control. These include both reward-shaping strategies and optimization objectives designed to encourage shortness without compromising accuracy. Methods like O1-Pruner \cite{luo2025o1prunerlengthharmonizingfinetuningo1like} use Proximal Policy Optimization (PPO) \cite{schulman2017proximalpolicyoptimizationalgorithms} with a length-harmonizing reward that penalizes deviations from a target length relative to a reference model. Another work \cite{yeo2025demystifyinglongchainofthoughtreasoning} applies a two-branch PPO strategy with cosine penalties based on the prediction's proximity to a maximum acceptable length. L1 \cite{aggarwal2025l1controllinglongreasoning}, whose aim is an exact token budget, impractical when problem difficulty is unknown, employs Grouped Reinforcement Policy Optimization (GRPO) \cite{shao2024deepseekmathpushinglimitsmathematical}, explicitly conditioning the model to "think for N tokens" and penalizing outputs exceeding learned optimal lengths. Simple Preference Optimization (SimPO)-based \cite{NEURIPS2024_e099c1c9} approaches like DAST \cite{shen2025dastdifficultyadaptiveslowthinkinglarge} use data constructed from user or synthetic feedback to learn token-length preferences, bypassing reliance on reference models. Other traditional \cite{han2025tokenbudgetawarellmreasoning} and policy gradient (PG) methods \cite{arora2025traininglanguagemodelsreason} have also been tried in this context, such as those including early fine-tuning with correctness-weighted penalties on output length.

Output length control, while previously a peripheral concern, has become central in reasoning LLMs \cite{sui2025stopoverthinkingsurveyefficient, fatemi2025concisereasoningreinforcementlearning}. Earlier works used architectural tweaks or static training signals, which lacked the flexibility to handle diverse inference scenarios. However, recent RL-based models \cite{butcher2024preciselengthcontrollarge, jie2023promptbasedlengthcontrolledgeneration, hou2025thinkprunepruninglongchainofthought} achieve better control by continuously adapting the reasoning length based on the learned reward signals. Yet many still rely on token count or heuristics as proxies for conciseness \cite{sui2025stopoverthinkingsurveyefficient}. Models such as C3oT \cite{Kang_Sun_Chen_Zou_2025} use LLMs as compressors to post-process reasoning traces, while continuous token methods like CCoT \cite{cheng2024compressedchainthoughtefficient} and latent-space approaches like COCONUT \cite{hao2024traininglargelanguagemodels} reduce token-level redundancy by replacing explicit steps abstract representations. Although effective at shortening outputs, these approaches operate post hoc or rely on static heuristics that limit adaptability across tasks.

To address these challenges, we propose a novel approach that uses an LLM as a judge to evaluate the conciseness of reasoning traces. While LLMs have been widely used as evaluators for factuality, summarization, coherence, and style \cite{NEURIPS2023_91f18a12, li2025generationjudgmentopportunitieschallenges, gu2025surveyllmasajudge}, leveraging them to assess reasoning conciseness remains unexplored. Our method introduces a semantically aware, dynamic reward function that goes beyond token count or static heuristics. This enables models to generate reasoning that is not only correct but also efficient and interpretable, bridging the gap between performance and inference cost.

\section{Method}
\label{sec:method}

We build on pre-trained multi-step reasoning models, improving them to generate correct answers using the minimum semantic effort through more concise outputs. Instead of penalizing length, we reward conciseness: the ability to justify an answer with the fewest necessary and non‑redundant steps. \textbf{Concise traces are often short, but short traces are not always concise, therefore the reward targets conciseness directly, and a shorter length appears only as a side effect.}

\subsection{Conciseness and Accuracy Rewards}
\label{subsec:scores}

Given an input prompt $x$ and a reasoning trace $y=(t_1,\ldots,t_{|y|})$ sampled from the model $p_\theta(\cdot \mid x)$, we compute a \textbf{conciseness score} $C(y)\in[0,1]$ by querying an external LLM judge $\mathcal{J}$, which receives the reasoning trace and evaluates it according to a custom system prompt (Appendix~\ref{appendix:conciseness_prompt}). The system prompt instructs the judge to assign a discrete conciseness score $s$ from 1 (overly verbose) to 10 (clear reasoning), disregarding correctness. The discrete score $s \in \{1,\dots,10\}$ is then normalized as: $C(y)\;=\;\frac{s}{10}\in[0,1]$.

\paragraph{Accuracy Signal.}
Let $y^\star(x)$ denote the ground‑truth answer to prompt $x$, and let $\mathrm{Ans}(y)$ be the final answer extracted from the reasoning trace (e.g. \textbackslash boxed\{$\mathrm{Ans}(y)$\}). The accuracy is then:

\begin{equation}
\label{eq:accuracy}
A(y,x)=
\begin{cases}
1, & \text{if }\ \mathrm{Ans}(y)=y^\star(x),\\[6pt]
0, & \text{otherwise.}
\end{cases}
\end{equation}

\subsection{Reward variants}
We employ two reward formulations:

\begin{enumerate}
    \item Pure conciseness  
          \begin{equation}
              R_{\mathrm{c}}(y,x)\;=\;C(y).
              \label{eq:reward_concise}
          \end{equation}

    \item Accuracy‑gated conciseness
          (used to avoid judge calls when a trace is already wrong)  
          \begin{equation}
              R_{\mathrm{ac}}(y,x)\;=\;A(y,x)\,\cdot\,C(y).
              \label{eq:reward_acc}
          \end{equation}
\end{enumerate}

$R_{\mathrm{ac}}$ is cheaper in terms of API calls because the judge is queried only when $A(y,x)=1$. On 1.5B models, $R_{\mathrm{ac}}$ costs $<$ \$9 per training run, and there is virtually no increase in training time. The cost of prompting the model stays constant relative to the size of the training data. The judge is not used during inference, so our method has no additional cost or overhead during deployment. In the experiments, \textbf{we denote the Accuracy-gated reward as ''ConciseRL'' and the Pure conciseness reward as ''ConciseRL (Separated)''}. ConciseRL uses only the gated conciseness as a reward, and the Separated variant includes both conciseness and accuracy as rewards during training.

\subsection{Optimization with PPO}
\label{subsec:grpo}

Let $\rho$ denote the distribution over prompts. Our objective is to maximize:

\begin{equation}
    \mathcal{J}(\theta)\;=\;\mathbb{E}_{x\sim\rho}\,\mathbb{E}_{y\sim p_\theta(\cdot|x)}\bigl[R(y,x)\bigr].
    \label{eq:objective}
\end{equation}

where $R$ is a reward variant. Since expectation $y$ is taken over sequences sampled autoregressively from $p_\theta$, $\mathcal{J}$ is not differentiable. Thus, we apply PPO \cite{shao2024deepseekmathpushinglimitsmathematical} with a
sequence‑level objective.

\subsection{Leave‑One‑Out Advantage}
Based on \cite{arora2025traininglanguagemodelsreason}, for each prompt we sample $n$ candidate traces $\{y_i\}_{i=1}^n$ and compute the trajectory return $R(y_i,x)$. The sequence‑level advantage is estimated with a Reinforced Leave‑One‑Out baseline:

\begin{equation}
    A(y_i,x)\;=\;R(y_i,x)-\frac{1}{n-1}\sum_{j\neq i} R(y_j,x).
    \label{eq:advantage}
\end{equation}

Using Equation~\eqref{eq:advantage}, the clipped PPO loss is:

\begin{equation}
    \mathcal{L}(\theta)
    \;=\;
    \mathbb{E}_{x,\,y}
    \Bigl[
        w_i
        \cdot
        \log p_\theta(y\,|\,x)
    \Bigr],
    \label{eq:grpo_loss}
\end{equation}

\begin{equation}
    w_i = \operatorname{clip}\left( \tfrac{p_\theta(y_i|x)}{p_{\theta_\text{old}}(y_i|x)}, 1{-}\epsilon, 1{+}\epsilon \right) A(y_i, x).
\end{equation}

where $\epsilon$ is the PPO clipping threshold and $p_{\theta_\text{old}}$ is the policy prior to the update.

\section{Experiments and Results}

\subsection{Experimental Setup}
\label{subsec:setup}

We begin with two publicly available reasoning models as our foundation: DeepSeek-R1-Distill-Qwen-1.5B \cite{deepseekai2025deepseekr1incentivizingreasoningcapability}, and STILL-3-1.5B-preview \cite{Slow_Thinking_with_LLMs_3_Preview, Slow_Thinking_with_LLMs_2}, conducting all ablations on the former with GPT-4.1 mini \cite{openaigpt41} as the LLM judge. The judge's system prompt is shown in Figure~\ref{fig:system_prompt}. Our RL setup follows Efficient Reasoning \cite{arora2025traininglanguagemodelsreason, numina_math_datasets}, including a learning rate of $5{\times}10^{-6}$, Kullback–Leibler (KL) divergence coefficient of $10^{-3}$, PPO clipping threshold of $0.2$, and a maximum context window of $32\text{K}$ tokens, $8$ rollouts per prompt, and a global batch of $128$ ($32$ prompts). All plots are smoothed using a Gaussian kernel of width $9$ for visual clarity. The Cosine Reward baseline uses hyperparameters from the Demystifying paper \cite{yeo2025demystifyinglongchainofthoughtreasoning}: $L_\text{max}{=}14336$, $rc_0{=}2.0$, $rc_L{=}1.0$, $rw_0{=}{-}10.0$, $rw_L{=}0.0$, and $r_\text{exceed}{=}{-}10.0$. The Efficient Reasoning baseline uses the same hyperparameters defined above, $\alpha \in {0.1, 0.2, 0.4, 0.6, 0.9}$, and is implemented in vLLM \cite{10.1145/3600006.3613165}. We do not compare with methods that are not open-source or not reproducible. For instance, the method~\cite{fatemi2025concisereasoningreinforcementlearning} is similar to ours, but their reward is based on shortening traces rather than conciseness, and their code is not publicly available. We also didn't include L1~\cite{aggarwal2025l1controllinglongreasoning} since it requires fixing the output length in advance, a fundamentally different setup that limits adaptability and can't learn dynamic conciseness. 

Training is conducted on 4$\times$NVIDIA A100 80GB GPUs. A complete run on the 1.5B model takes $\sim$ $20$ GPU-hours. We evaluate model performance across multiple benchmarks: GSM8K \cite{cobbe2021trainingverifierssolvemath}, MATH500 \cite{hendrycks2021measuringmathematicalproblemsolving}, TheoremQA \cite{chenetal2023theoremqa}, GPQA-main \cite{rein2024gpqa}, and MMLU-Pro-1k \cite{NEURIPS2024_ad236edc}. 

\subsection{Dynamic vs Static Rewards}

\begin{table*}[ht]
  \centering
  \resizebox{\textwidth}{!}{%
  \begin{tabular}{l*{6}{cc}}
    \toprule
    & \multicolumn{2}{c}{\textbf{GSM8K}} 
    & \multicolumn{2}{c}{\textbf{MATH500}} 
    & \multicolumn{2}{c}{\textbf{MMLU‑Pro-1k}} 
    & \multicolumn{2}{c}{\textbf{GPQA-main}} 
    & \multicolumn{2}{c}{\textbf{TheoremQA}} 
    & \multicolumn{2}{c}{\textbf{Average}} \\
    \cmidrule(lr){2-3}\cmidrule(lr){4-5}\cmidrule(lr){6-7}%
    \cmidrule(lr){8-9}\cmidrule(lr){10-11}\cmidrule(lr){12-13}
    \textbf{Technique} &
    \textbf{Acc.} & \textbf{Len.(\%)} &
    \textbf{Acc.} & \textbf{Len.(\%)} &
    \textbf{Acc.} & \textbf{Len.(\%)} &
    \textbf{Acc.} & \textbf{Len.(\%)} &
    \textbf{Acc.} & \textbf{Len.(\%)} &
    \textbf{Acc.} & \textbf{Len.(\%)} \\
    \midrule
    \textbf{ConciseRL (Separated)}              & \textbf{72.5} & 16.4 & \textbf{68.6} & 16.3 & \textbf{19.2} & 19.5 & \textbf{31.0} & 19.7 & \textbf{28.5} & 8.0 & \textbf{44.0} & 16.0 \\
    Eff. Reasoning $\alpha{=}0.9$    & 16.1 & 1.2 & 33.6 & 21.0 & 18.2 & 11.6 & 27.9 & 26.2 & 20.4 & 9.7 & 23.2 & 13.9  \\
    \midrule
    \textbf{ConciseRL}                       & \textbf{80.9} & 35.8 & \textbf{78.0} & 30.8 & 24.5 & 38.5 & 30.4 & 45.9 & 32.5 & 21.0 & \textbf{49.3} & 34.4 \\
    Eff. Reasoning $\alpha{=}0.6$ & 55.9 & 23.9 & 64.4 & 26.2 & \textbf{27.8} & 44.2 & \textbf{35.0} & 55.0 & \textbf{34.3} & 23.5 & 43.5 & 34.6 \\
    \midrule
    Eff. Reasoning $\alpha{=}0.4$& 74.7 & 41.9 & 76.8 & 32.4 & 21.8 & 45.2 & 31.5 & 52.1 & 33.3 & 31.2 & 47.6 & 40.6 \\
    Eff. Reasoning $\alpha{=}0.2$& 79.8 & 43.4 & 79.6 & 32.8 & 22.2 & 38.6 & 29.9 & 62.0 & 34.9 & 41.7 & 49.3 & 43.7 \\
    Eff. Reasoning $\alpha{=}0.1$& \textbf{82.5} & 63.7 & 78.4 & 43.9 & 21.6 & 42.6 & 32.1 & 56.9 & 34.5 & 43.0 & 49.8 & 50.0 \\
    Cosine Reward               & 80.4 & 310.8 & 77.0 & 79.9 & 19.8 & 93.8 & 32.1 & 91.3 & 29.4 & 93.2 & 47.7 & 133.8 \\
    DeepScaleR                    & 80.7 & 332.7  & 82.6 & 60.0 & 31.5 & 67.7 & 32.6 & 71.3 & 34.8 & 57.0 & 52.4 & 117.7 \\
    Full Reasoning                     & 76.3 & 100.0 & 71.4 & 100.0 & 25.8 & 100.0 & 26.6 & 100.0 & 26.3 & 100.0 & 45.3 & 100.0 \\
    \bottomrule
  \end{tabular}%
  }
  \caption{Comparison of accuracy (higher is better) and token length (as \% of Full Reasoning; lower is better) across datasets for DeepSeek-R1-Distill-Qwen-1.5B. We group our methods with baselines that reach similar average reasoning-trace length, and \textbf{bold the best accuracy within each such group}. This highlights that our methods achieve stronger performance at similar trace length.}
  \label{tab:exact_results}
\end{table*}

\begin{table*}[ht]
  \centering
  \resizebox{\textwidth}{!}{%
  \begin{tabular}{l*{6}{cc}}
    \toprule
    & \multicolumn{2}{c}{\textbf{GSM8K}} 
    & \multicolumn{2}{c}{\textbf{MATH500}} 
    & \multicolumn{2}{c}{\textbf{MMLU-Pro-1k}} 
    & \multicolumn{2}{c}{\textbf{GPQA-main}} 
    & \multicolumn{2}{c}{\textbf{TheoremQA}} 
    & \multicolumn{2}{c}{\textbf{Average}} \\
    \cmidrule(lr){2-3}\cmidrule(lr){4-5}\cmidrule(lr){6-7}%
    \cmidrule(lr){8-9}\cmidrule(lr){10-11}\cmidrule(lr){12-13}
    \textbf{Technique} &
    \textbf{Acc.} & \textbf{Len.(\%)} &
    \textbf{Acc.} & \textbf{Len.(\%)} &
    \textbf{Acc.} & \textbf{Len.(\%)} &
    \textbf{Acc.} & \textbf{Len.(\%)} &
    \textbf{Acc.} & \textbf{Len.(\%)} &
    \textbf{Acc.} & \textbf{Len.(\%)} \\
    \midrule
    \textbf{ConciseRL}                  & \textbf{78.5} & 11.0 & \textbf{77.2} & 33.7 & \textbf{19.9} & 32.3 & 28.3 & 34.3 & 29.6 & 20.6 & \textbf{46.7} & 26.4 \\
    Eff. Reasoning $\alpha{=}0.4$       & 60.9 & 4.9 & 68.0 & 27.8 & 18.8 & 31.0 & \textbf{31.5} & 36.1 & \textbf{33.1} & 20.1 & 42.5 & 24.0 \\
    \midrule
    Eff. Reasoning $\alpha{=}0.9$       & 14.6 & 0.4 & 18.8 & 2.0 & 14.6 & 0.1 & 27.7 & 0.2 & 16.5 & 0.1 & 18.4 & 0.6 \\
    Eff. Reasoning $\alpha{=}0.6$       & 81.3 & 17.9 & 77.6 & 44.1 & 24.3 & 56.3 & 31.2 & 53.9 & 34.1 & 45.5 & 49.7 & 43.5 \\
    Eff. Reasoning $\alpha{=}0.2$       & 81.3 & 17.9 & 77.6 & 44.1 & 24.3 & 56.3 & 31.2 & 53.9 & 34.1 & 45.5 & 49.7 & 43.5 \\
    Eff. Reasoning $\alpha{=}0.1$       & 85.0 & 34.6 & 82.0 & 56.6 & 20.4 & 68.5 & 30.4 & 82.2 & 36.1 & 52.6 & 50.8 & 58.9 \\
    Cosine Reward                       & 43.6 & 598.7 & 44.4 & 298.8 & 17.8 & 216.2 & 28.8 & 138.5 & 13.6 & 199.9 & 29.6 & 290.4 \\
    DeepScaleR                    & 80.7 & 138.0  & 82.6 & 82.5 & 31.5 & 75.7 & 32.6 & 72.1 & 34.8 & 80.6 & 52.4 & 89.8 \\
    Full Reasoning                      & 83.1 & 100.0 & 78.8 & 100.0 & 32.0 & 100.0 & 31.0 & 100.0 & 29.8 & 100.0 & 50.9 & 100.0 \\
    \bottomrule
  \end{tabular}%
  }
  \caption{Comparison of accuracy (higher is better) and token length (as \% of Full Reasoning; lower is better) across datasets for STILL-3-1.5B-preview. We group our methods with baselines that reach similar average reasoning-trace length, and \textbf{bold the best accuracy within each such group}. This highlights that our methods achieve stronger performance at similar trace length.}
  \label{tab:exact_results_still_exp}
\end{table*}

We compare our method to strong baselines across five benchmarks. Baselines include Efficient Reasoning~\cite{arora2025traininglanguagemodelsreason}, which introduces a length penalty modulated by a hyperparameter $\alpha \in [0, 1)$. Higher values of $\alpha$ increase the penalty for longer generations, leading to shorter reasoning but compromising accuracy. We also compare against Cosine Reward~\cite{yeo2025demystifyinglongchainofthoughtreasoning}, which assigns rewards using a cosine function that favors correct answers, modulating reward magnitude based on CoT length, and DeepScaleR~\cite{deepscaler2025}. The full reasoning baseline corresponds to the model's default generation without additional training to shorten the reasoning traces. The "Separated" baseline is missing from Table~\ref{tab:exact_results_still_exp} due to substantial training costs given our budget constraints. Unlike these methods, our approach is hyperparameter-free, making it more robust and easier to deploy. 

\begin{figure*}[ht]
    \centering
    \includegraphics[width=\linewidth]{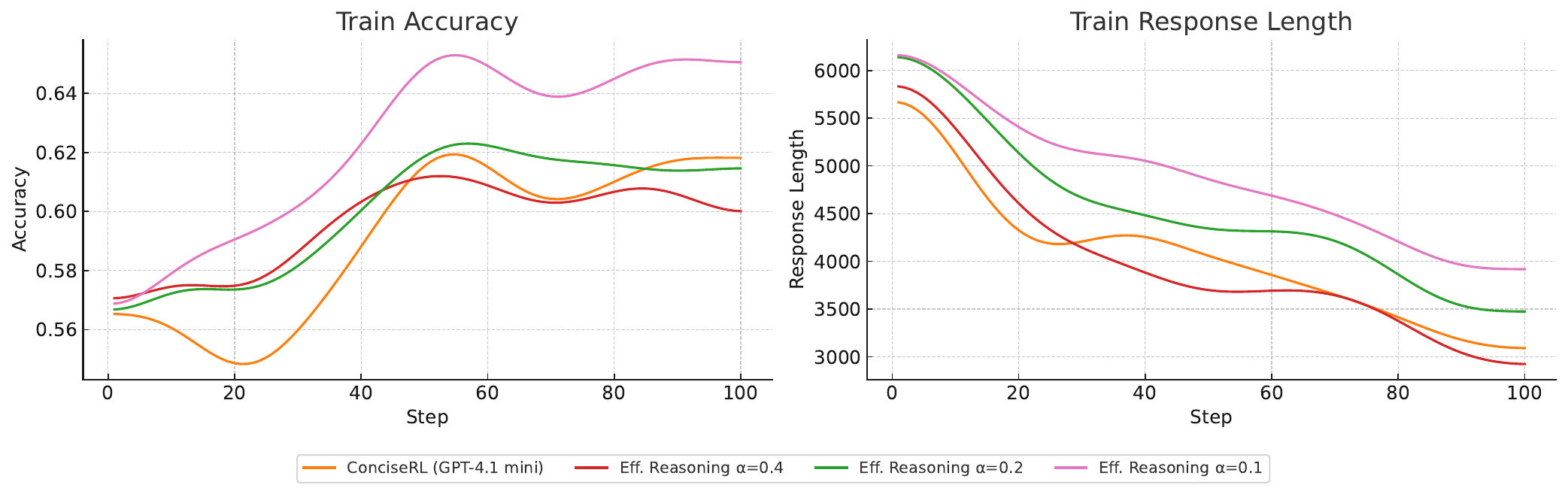}
    \caption{Training metrics across steps using DeepSeek-R1-Distill-Qwen-1.5B as the base model. The Y-axes show accuracy (higher is better) and response length in tokens (lower is better).}
    \label{fig:training_metrics_1}
\end{figure*}

As shown in Table~\ref{tab:exact_results}, we group methods with similar average reasoning lengths and compare their accuracy to assess efficiency–accuracy trade-offs. Our approach, ConciseRL (Separated), achieves significantly higher accuracy compared to Efficient Reasoning with $\alpha{=}0.9$, with an average improvement of 20.8\% at comparable reasoning lengths, and up to 56.4\% on benchmarks like MATH500 and GSM8K. ConciseRL outperforms Efficient Reasoning $\alpha{=}0.6$ by 5.8\% accuracy on average. Compared to the full reasoning model, ConciseRL reduces token usage by 65.6\% while boosting accuracy by 4\% on average. The separated variant is even more concise: it maintains 44.0\% average accuracy while using only 16\% of the full reasoning trace. While Efficient Reasoning with $\alpha{=}0.6$ achieves similar token savings (34.6\%) to ConciseRL, it severely underperforms in accuracy, especially on GSM8K and MATH500, with an overall drop to 43.5\% average accuracy compared to our 49.3\%. This underscores the difficulty of tuning static penalties and the benefit of our dynamic conciseness reward.

This pattern also appears in Table~\ref{tab:exact_results_still_exp}, evaluating STILL-3-1.5B-preview. ConciseRL delivers the best overall balance, consistently maintaining strong average accuracy with low reasoning length. At comparable length, our method outperforms Efficient Reasoning $\alpha{=}0.4$ by 4.2\% in average accuracy. In contrast, Efficient Reasoning displays unstable performance across $\alpha$ values: while $\alpha{=}0.1$, $\alpha{=}0.2$, and $\alpha{=}0.6$ retain high accuracy but produce moderately long traces, aggressive settings like $\alpha{=}0.9$ sharply reduce length but cause severe performance drops. Hence, the Efficient Reasoning method displays noticeable hyperparameter sensitivity and requires considerable fine-tuning to achieve favorable performance-length trade-offs. Our ConciseRL method avoids this issue by not relying on additional hyperparameters or manual tuning. For examples of reasoning traces from different methods, see Section~\ref{appendix:reasoning_trace_examples}.

During training, our model follows a similar accuracy trajectory as Efficient Reasoning $\alpha$ = 0.2 but converges to response lengths comparable to $\alpha$ = 0.4 as can be seen in Figure~\ref{fig:training_metrics_1}, thus achieving the benefits of both ends of the trade-off curve. Additionally, Figure~\ref{fig:full_training} shows that our separated variant achieves the shortest reasoning traces among all methods, with an accuracy above the $\alpha = 0.6$ baseline. The conciseness signal from our semantic reward enables this fine-grained control, in contrast to static heuristics that uniformly penalize length. This adaptivity is also reflected in the performance across the other datasets: our method achieves the shortest reasoning traces in all benchmarks among methods with similar accuracy. These results indicate that conciseness-aware optimization offers a more efficient path to high-quality reasoning than static length constraints or cosine penalties. We also perform a KL divergence analysis (Appendix~\ref{appendix:KL_divergence}), where our method shows more stable and efficient policy updates compared to other static baselines.

\subsection{Judge Model Comparison}

\begin{table*}[ht]
  \centering
  \resizebox{\textwidth}{!}{%
  \begin{tabular}{l*{6}{cc}}
    \toprule
    & \multicolumn{2}{c}{\textbf{GSM8K}} 
    & \multicolumn{2}{c}{\textbf{MATH500}} 
    & \multicolumn{2}{c}{\textbf{MMLU‑Pro-1k}} 
    & \multicolumn{2}{c}{\textbf{GPQA-main}} 
    & \multicolumn{2}{c}{\textbf{TheoremQA}} 
    & \multicolumn{2}{c}{\textbf{Average}} \\
    \cmidrule(lr){2-3}\cmidrule(lr){4-5}\cmidrule(lr){6-7}%
    \cmidrule(lr){8-9}\cmidrule(lr){10-11}\cmidrule(lr){12-13}
    \textbf{Technique} &
    \textbf{Acc.} & \textbf{Len.(\%)} &
    \textbf{Acc.} & \textbf{Len.(\%)} &
    \textbf{Acc.} & \textbf{Len.(\%)} &
    \textbf{Acc.} & \textbf{Len.(\%)} &
    \textbf{Acc.} & \textbf{Len.(\%)} &
    \textbf{Acc.} & \textbf{Len.(\%)} \\
    \midrule
    ConciseRL (GPT‑4.1 mini)         & 80.9 & 35.8 & 78.0 & 30.8 & 24.5 & 38.5 & 30.4 & 45.9 & 32.5 & 21.0 & 49.3 & 34.4 \\
    ConciseRL (GPT‑4o mini)          & 82.8 & 34.0 & 77.4 & 28.5 & 19.1 & 28.5 & 28.6 & 36.4 & 34.8 & 17.9 & 48.5 & 29.1 \\
    ConciseRL (GPT‑4.1 nano)         & 83.6 &190.0 & 75.6 & 76.6 & 17.5 & 58.7 & 30.6 & 65.8 & 29.3 & 85.8 & 47.3 & 95.4 \\
    Full Reasoning                   & 76.3 &100.0 & 71.4 &100.0 & 25.8 &100.0 & 26.6 &100.0 & 26.3 &100.0 & 45.3 &100.0 \\
    \bottomrule
  \end{tabular}%
  }
  \caption{Comparison of accuracy (higher is better) and token length (as \% of Full Reasoning; lower is better) across datasets for DeepSeek-R1-Distill-Qwen-1.5B when using different judge models.}
  \label{tab:gpt_comparison}
\end{table*}

The quality of the LLM judge used to assess conciseness significantly influences the final model behavior. Figure~\ref{fig:training_metrics_2} (appendix) shows that a more capable judge enables more effective optimization. GPT-4.1 mini \cite{openaigpt41} and GPT-4o mini \cite{openaigpt4o} yield models that significantly reduce response length during training while maintaining or improving accuracy. In contrast, GPT-4.1 nano \cite{openaigpt41}, despite producing slightly higher training accuracy in the end, does not meaningfully shorten the reasoning traces, likely because the score has more "noise". We highlight that the accuracy curve is not necessarily indicative of final model performance: accuracy rises early as the model learns the dataset but later decreases when the reward begins favoring conciseness more strongly and the traces become shorter.  

Tables~\ref{tab:exact_results} and \ref{tab:gpt_comparison} show the efficiency–accuracy trade-offs introduced by each judge. GPT-4o mini, in particular, achieves the highest average accuracy (48.5\%) with only 29.1\% of the tokens. GPT-4.1 mini delivers a strong efficiency–accuracy trade-off, matching the average accuracy of Efficient Reasoning at $\alpha$ = 0.2 (49.3\%) while producing 65.6\% shorter reasoning traces. On the other hand, GPT-4.1 nano offers limited trace compression (95.4\% of full reasoning) despite comparable accuracy (47.3\%), validating the earlier observation that noisier reward signals make optimization harder.

\subsection{The Behavior of the Rewards}

\begin{figure}[ht]
    \begin{subfigure}[b]{0.49\textwidth}
        \centering
        \includegraphics[width=\linewidth]{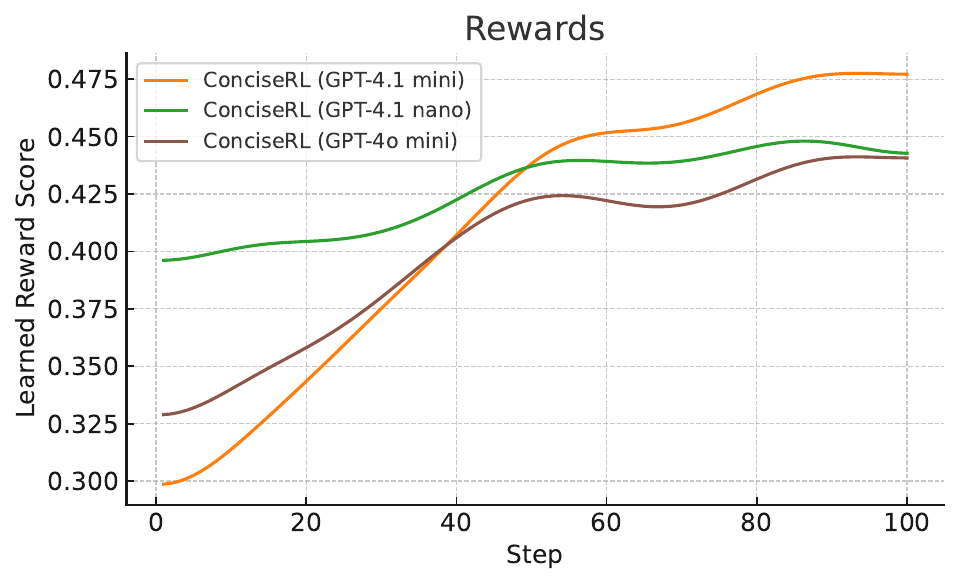}
        \caption{GPTs Rewards}
        \label{fig:train_response3}
    \end{subfigure}
    \hfill
    \begin{subfigure}[b]{0.49\textwidth}
        \centering
        \includegraphics[width=\linewidth]{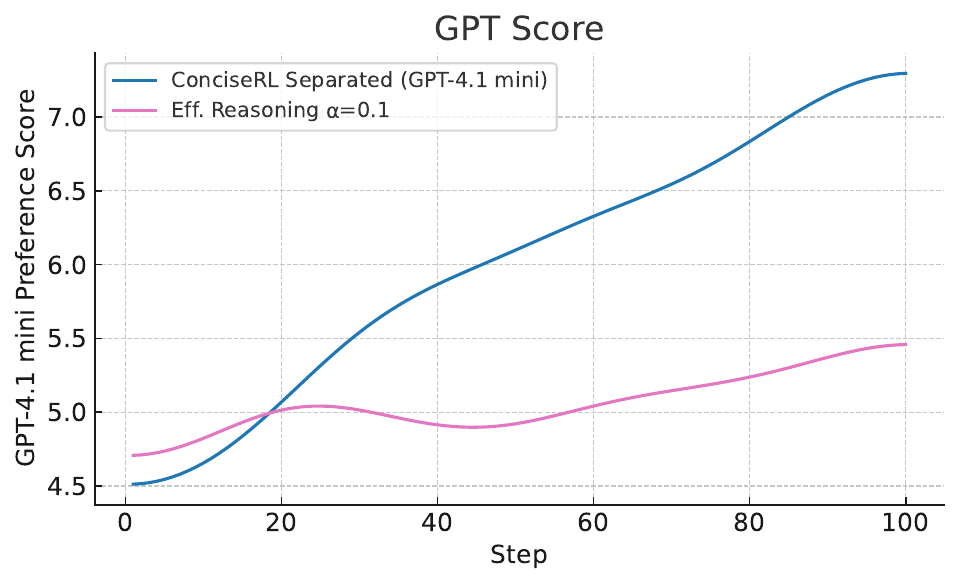}
        \caption{GPT Score}
        \label{fig:train_response4}
    \end{subfigure}

    \caption{Training metrics across steps using different models as the judge. The Y-axes show reward values and conciseness scores assigned by the judge.}
    \label{fig:training_metrics_3}
\end{figure}

In Figure~\ref{fig:train_response3}, we compare reward values across different judges. Despite starting with the lowest average reward, the model trained with GPT‑4.1 mini ends with the highest, indicating that the reward, though initially strict, was more learnable and resulted in effective policy improvement. This contrasts with GPT‑4.1 nano, which yields higher early rewards but fails to produce meaningful shortening, likely due to its noisy and less discriminative signal. Notably, these values represent final reward outputs: a score of zero is assigned if the model's answer is incorrect (Equation \ref{eq:accuracy}), meaning the curves are also influenced by accuracy. 

In Figure~\ref{fig:train_response4}, we isolate conciseness from correctness using our separated reward configuration. The resulting GPT score, evaluated solely on reasoning quality, improves almost linearly from roughly 4.5 to 7.5 out of 10 over the course of training. Compared to Efficient Reasoning $\alpha{=}0.1$, we observe that although it successfully reduces trace length to nearly half (Table~\ref{tab:exact_results_still_exp}), its corresponding conciseness score only increases from 4.75 to 5.5. In contrast, our method boosts the score from 4.5 to 7.1. This shows that shorter traces are not necessarily more concise, reinforcing our argument that brevity should be a learned outcome of semantic compactness, not a heuristic target. 

\subsection{Performance by Problem Difficulty}

Figure~\ref{fig:histogram} and Table~\ref{tab:math_difficulty_levels} (appendix) show MATH500 performance across five difficulty levels. ConciseRL (Separated) achieves up to 31$\times$ fewer tokens than the Full Reasoning baseline on Level 1 (118 vs. 3686 tokens) while improving accuracy by ~7\%. For the hardest problems (Level 5), our method outperforms the Full Reasoning baseline using 3.6$\times$ fewer tokens (4431 vs. 16156 tokens). Our method exhibits a strong correlation between problem difficulty and reasoning length: easier problems are solved with fewer tokens, while harder ones elicit longer reasoning. This adaptive behavior aligns with human intuition and is a desirable trait in reasoning models. In contrast, other techniques like Efficient Reasoning ($\alpha=0.1$), Cosine Reward, or Full Reasoning either under-adapt (e.g., generating long traces even for trivial problems) or show no clear correlation. 

Furthermore, as shown in Table~\ref{tab:exact_results}, on the TheoremQA benchmark, our method achieves 2.2\% higher accuracy while using 12.5$\times$ fewer tokens compared to the Full Reasoning model. These results suggest that our semantically guided reward encourages selective elaboration, producing concise reasoning where appropriate and expanding only when necessary. Also, our method produces more explainable and compressed reasoning traces, see Appendix~\ref{appendix:reasoning_trace_examples} for examples and analysis.

\section{Conclusion}

In this work, we introduce a novel, hyperparameter-free conciseness score used as a reward signal within any reinforcement learning framework and evaluated by an LLM judge to encourage models to generate correct and concise reasoning traces. Our approach improves semantic density without sacrificing accuracy, yielding strong efficiency–accuracy trade-offs. At equal token length, our traces are more informative, interpretable, and explainable than those from static length-penalized baselines due to fewer filler steps and tighter logic. Overall, our results highlight the value of semantic rewards and LLM-based judges for guiding concise and correct reasoning.

In future work, we plan to evaluate our method across a wider range of model sizes to better understand how conciseness rewards scale. Additionally, while we hypothesize that using more advanced judges like GPT-4o or GPT-4.5 could further enhance performance, we could not evaluate this due to our budget limitations and API-associated costs. We are also interested in combining our conciseness optimization with orthogonal methods, such as structure-aware search or explicit length conditioning, to enhance reasoning quality and efficiency. Unlike other methods, ours is orthogonal and integrates well with approaches like L1 \cite{aggarwal2025l1controllinglongreasoning} (see \ref{appendix:future_work}).

\section*{Limitations}

Despite the strong empirical results, our study has several limitations. Due to limited hardware resources, including access to only 4$\times$NVIDIA A100 80GB GPUs, we could not evaluate larger models, test our approach at scale, or fully compare against all relevant baselines. Training with reinforcement learning is particularly resource-intensive, which further constrained the scope of our experiments. We were also only able to partially explore combinations of our method with other recent techniques. Finally, we restricted our use of LLMs-as-judges to smaller models because of the cost of using larger models, specifically, the higher cost for long reasoning traces. 

\section*{Ethical Considerations}

Regarding impact, we believe that our method has positive implications for reducing the computational cost of reasoning at inference time, particularly for models deployed in environments with constrained compute budgets or latency requirements. By shifting the optimization target from token length to conciseness, we aim to improve efficiency and interpretability without sacrificing correctness. 

However, we acknowledge several potential ethical risks. Using LLMs as evaluators introduces potential biases, which may reflect or amplify artifacts from their training data. If the LLM judge favors certain styles of explanation over others, this may steer the trained model toward narrow patterns of reasoning. Additionally, there is a risk that optimizing for conciseness could suppress contextually relevant reasoning steps, leading to overly truncated outputs. We mitigate these concerns by evaluating correctness alongside conciseness and analyzing behavior across problem difficulties and prompt variants. We also ensure that the LLM judge is only used during training, not at inference time, thus avoiding deployment dependencies.

Regarding safeguards, since our method builds upon publicly available pre-trained models without introducing new high-risk capabilities, we assess the risk of misuse to be low. Nevertheless, to support responsible usage, we commit to open-source the code, model weights, and datasets under a research license with clear documentation.



\bibliography{custom}

\appendix

\section{Appendix}

\subsection{Performance by Problem Difficulty}

To better understand how different methods adapt to problem complexity, we report accuracy and reasoning trace length across five difficulty levels in the MATH500 dataset. Table~\ref{tab:math_difficulty_levels} presents a breakdown of performance, highlighting that ConciseRL (Separated) achieves substantial token reductions—up to 31× on easy problems—while maintaining or improving accuracy. This demonstrates the model’s ability to dynamically adjust reasoning depth based on problem difficulty, a desirable trait not observed in baseline methods.

\begin{table*}[ht]
  \centering
  \resizebox{\textwidth}{!}{%
  \begin{tabular}{l*{6}{cc}}
    \toprule
    \textbf{Technique} &
    \multicolumn{2}{c}{\textbf{Level 1}} &
    \multicolumn{2}{c}{\textbf{Level 2}} &
    \multicolumn{2}{c}{\textbf{Level 3}} &
    \multicolumn{2}{c}{\textbf{Level 4}} &
    \multicolumn{2}{c}{\textbf{Level 5}} &
    \multicolumn{2}{c}{\textbf{Average}} \\
    \cmidrule(lr){2-3} \cmidrule(lr){4-5} \cmidrule(lr){6-7} \cmidrule(lr){8-9} \cmidrule(lr){10-11} \cmidrule(lr){12-13}
    & \textbf{Acc.} & \textbf{Len.}
    & \textbf{Acc.} & \textbf{Len.}
    & \textbf{Acc.} & \textbf{Len.}
    & \textbf{Acc.} & \textbf{Len.}
    & \textbf{Acc.} & \textbf{Len.}
    & \textbf{Acc.} & \textbf{Len.} \\
    \midrule
    ConciseRL (Separated) & 95.35 & \textbf{118.40} & 77.78 & \textbf{159.95} & 80.95 & \textbf{476.69} & 64.84 & \textbf{1470.79} & 47.78 & \textbf{4431.25} & 73.34 & \textbf{1331.42} \\
    ConciseRL           & 95.35 & 265.58 & 87.77 & 453.94 & 87.62 & 2073.16 & 75.78 & 2738.97 & 57.46 & 8029.31 & 80.80 & 2712.19 \\
    Eff. Reasoning $\alpha{=}0.1$ & 93.02 & 1285.63 & 88.88 & 1102.38 & 88.57 & 3355.63 & 74.22 & 4536.74 & 62.69 & 9004.73 & 81.48 & 3857.02 \\
    Full Reasoning      & 88.37 & 3686.48 & 85.56 & 5100.22 & 77.14 & 9949.32 & 73.44 & 10890.66 & 50.00 & 16156.81 & 74.90 & 9156.70 \\
    \bottomrule
  \end{tabular}%
  }
    \caption{Comparison of accuracy (higher is better) and token length (lower is better) by difficulty level on MATH500 \cite{hendrycks2021measuringmathematicalproblemsolving}, using DeepSeek-R1-Distill-Qwen-1.5B \cite{deepseekai2025deepseekr1incentivizingreasoningcapability}. For ConciseRL, we use GPT-4.1 mini \cite{openaigpt41} as the conciseness judge. \textbf{Bold} indicates the shortest reasoning traces.}
      \label{tab:math_difficulty_levels}
\end{table*}

\subsection{Extended Results}
\label{appendix:results_token_lengths}

\begin{figure*}[ht]
    \centering
    \includegraphics[width=\linewidth]{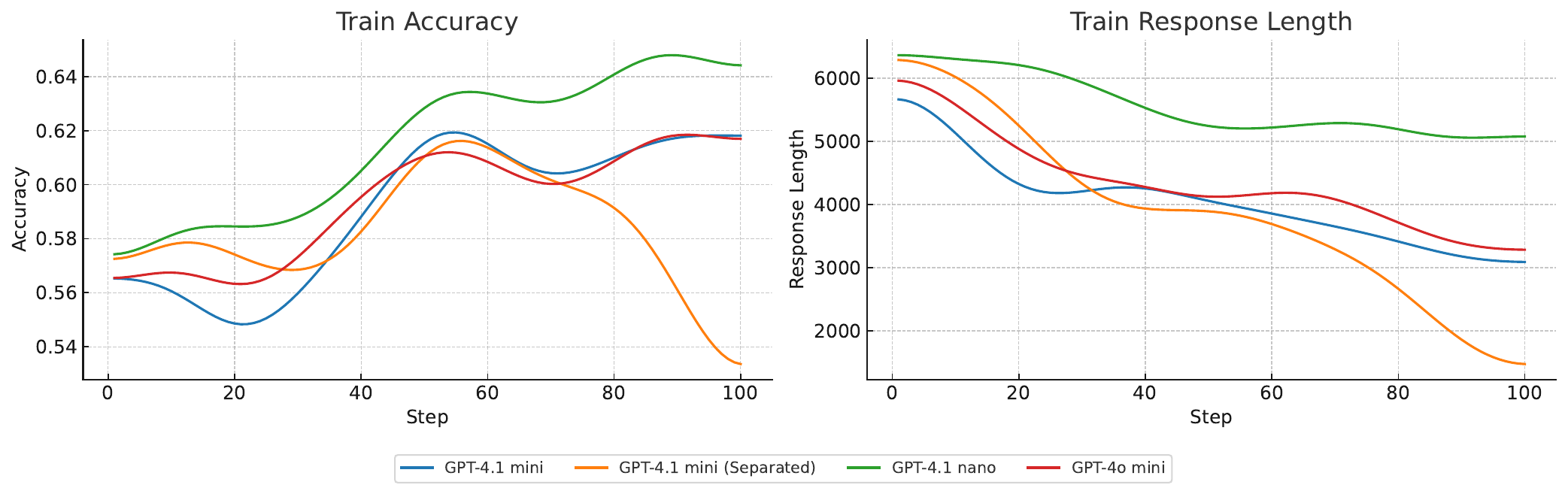}
    \caption{Training metrics across steps using DeepSeek-R1-Distill-Qwen-1.5B \cite{deepseekai2025deepseekr1incentivizingreasoningcapability} as the base model and different models as the judge. The Y-axes show accuracy (higher is better; left) and response length in tokens (lower is better; right). The X-axis in both cases shows the training step.}
    \label{fig:training_metrics_2}
\end{figure*}

As a supplement to our main findings, we provide additional results that reinforce the effectiveness and generality of our approach. Figure~\ref{fig:training_metrics_2} compares training dynamics under different LLM judges, showing that stronger judges like GPT-4.1 mini and GPT-4o mini enable more effective reasoning compression while preserving or improving accuracy. In contrast, GPT-4.1 nano yields higher variance and fails to consistently reduce reasoning length. These trends confirm that the reliability of the conciseness signal strongly depends on the evaluator’s capability \cite{openaigpt41}.

\begin{figure*}[ht]
    \centering
    \includegraphics[width=\linewidth]{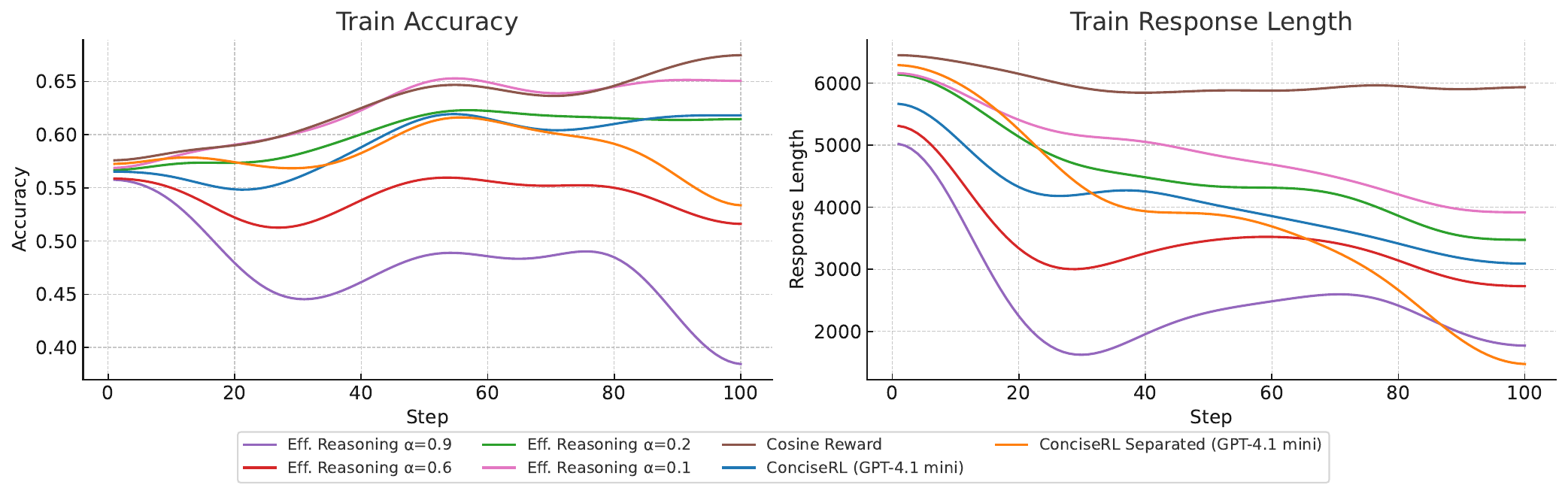}
    \caption{Training metrics across steps using DeepSeek-R1-Distill-Qwen-1.5B \cite{deepseekai2025deepseekr1incentivizingreasoningcapability} as the base model and GPT-4.1 mini \cite{openaigpt41} as the judge. The Y-axes show accuracy (higher is better; left) and response length in tokens (lower is better; right). The X-axis in both cases shows the training step.}
    \label{fig:full_training}
\end{figure*}

Figure~\ref{fig:full_training} extends out comparison with static baselines such as Efficient Reasoning \cite{deepseekai2025deepseekr1incentivizingreasoningcapability} display a trade-off between brevity and correctness, with aggressive penalties reducing token usage but at the cost of sharp accuracy drops. Our method, ConciseRL, follows a more desirable trajectory: it achieves the conciseness of the most efficient baselines while maintaining high accuracy, effectively tracing out the Pareto frontier.

Table~\ref{tab:exact_results_3} presents raw token counts on the DeepSeek-R1-Distill-Qwen-1.5B model across all benchmarks. The separated reward variant (ConciseRL Separated) yields the most compact reasoning traces, while ConciseRL strikes a strong efficiency–accuracy balance. Notably, both outperform static methods like Cosine Reward and Efficient Reasoning across all datasets.

Finally, Table~\ref{tab:exact_results_still_exp_raw} replicates this comparison on the STILL-3-1.5B-preview model \cite{Slow_Thinking_with_LLMs_3_Preview, Slow_Thinking_with_LLMs_2}, further supporting the generalizability of our framework. The same trends persist: semantic conciseness rewards outperform length-based baselines, both in efficiency and accuracy. These results confirm that our approach is robust across architectures, training signals, and judge variants.

\begin{table*}[ht]
  \centering
  \resizebox{\textwidth}{!}{%
  \begin{tabular}{l*{7}{cc}}
    \toprule
    & \multicolumn{2}{c}{\textbf{GSM8K}}
    & \multicolumn{2}{c}{\textbf{MATH500}}
    & \multicolumn{2}{c}{\textbf{MMLU-Pro-1k}}
    & \multicolumn{2}{c}{\textbf{GPQA-main}}
    & \multicolumn{2}{c}{\textbf{TheoremQA}}
    & \multicolumn{2}{c}{\textbf{Average}} \\
    \cmidrule(lr){2-3}\cmidrule(lr){4-5}\cmidrule(lr){6-7}\cmidrule(lr){8-9}\cmidrule(lr){10-11}\cmidrule(lr){12-13}
    \textbf{Technique} &
    \textbf{Acc.} & \textbf{Len.} &
    \textbf{Acc.} & \textbf{Len.} &
    \textbf{Acc.} & \textbf{Len.} &
    \textbf{Acc.} & \textbf{Len.} &
    \textbf{Acc.} & \textbf{Len.} &
    \textbf{Acc.} & \textbf{Len.} \\
    \midrule
    ConciseRL (Separated)             & 72.5 & 248   & 68.6 & 1703  & 19.2 & 2904  & 31.0 & 4913  & 28.5 & 1534  & 43.96 & 2260 \\
    ConciseRL                         & 80.9 & 543   & 78.0 & 3221  & 24.5 & 5732  & 30.4 & 11429 & 32.5 & 4034  & 49.26 & 4992 \\
    ConciseRL (GPT-4o mini)           & 82.8 & 515   & 77.4 & 2971  & 19.1 & 4240  & 28.6 & 9045  & 34.8 & 3439  & 48.54 & 4042 \\
    ConciseRL (GPT-4.1 nano)          & 83.6 & 2880  & 75.6 & 7994  & 17.5 & 8738  & 30.6 & 16363 & 29.3 & 16510 & 47.32 & 10497 \\
    Eff. Reasoning $\alpha{=}0.9$     & 16.1 & 19    & 33.6 & 2195  & 18.2 & 1723  & 27.9 & 6509  & 20.4 & 1872  & 23.24 & 2463 \\
    Eff. Reasoning $\alpha{=}0.6$     & 55.9 & 362   & 64.4 & 2735  & 27.8 & 6583  & 35.0 & 13675 & 34.3 & 4533  & 43.48 & 5578 \\
    Eff. Reasoning $\alpha{=}0.4$     & 74.7 & 635   & 76.8 & 3382  & 21.8 & 6730  & 31.5 & 12973 & 33.3 & 6009  & 47.62 & 5946 \\
    Eff. Reasoning $\alpha{=}0.2$     & 79.8 & 658   & 79.6 & 3428  & 22.2 & 5746  & 29.9 & 15415 & 34.9 & 8029  & 49.28 & 6655 \\
    Eff. Reasoning $\alpha{=}0.1$     & 82.5 & 966   & 78.4 & 4588  & 21.6 & 6347  & 32.1 & 14157 & 34.5 & 8271  & 49.82 & 6866 \\
    Cosine Reward                     & 80.4 & 4711  & 77.0 & 8348  & 19.8 & 13962 & 32.1 & 22708 & 29.4 & 17933 & 47.74 & 13532 \\
    DeepScaleR                    & 80.7 & 5044  & 82.6 & 6160 & 31.5 & 10071 & 32.6 & 17742 & 34.8 & 10970 & 52.4 & 9997.4 \\
    Full Reasoning                    & 76.3 & 1516  & 71.4 & 10442 & 25.8 & 14886 & 26.6 & 24880 & 26.3 & 19251 & 45.28 & 14195 \\
    \bottomrule
  \end{tabular}%
  }
  \caption{Comparison of accuracy (higher is better) and token length (lower is better) across datasets for DeepSeek-R1-Distill-Qwen-1.5B \cite{deepseekai2025deepseekr1incentivizingreasoningcapability}. For ConciseRL, we use GPT-4.1 mini \cite{openaigpt41} as the conciseness judge.}

  \label{tab:exact_results_3}
\end{table*}

\begin{table*}[ht]
  \centering
  \resizebox{\textwidth}{!}{%
  \begin{tabular}{l*{6}{cc}}
    \toprule
    & \multicolumn{2}{c}{\textbf{GSM8K}} 
    & \multicolumn{2}{c}{\textbf{MATH500}} 
    & \multicolumn{2}{c}{\textbf{MMLU-Pro-1k}} 
    & \multicolumn{2}{c}{\textbf{GPQA-main}} 
    & \multicolumn{2}{c}{\textbf{TheoremQA}} 
    & \multicolumn{2}{c}{\textbf{Average}} \\
    \cmidrule(lr){2-3}\cmidrule(lr){4-5}\cmidrule(lr){6-7}%
    \cmidrule(lr){8-9}\cmidrule(lr){10-11}\cmidrule(lr){12-13}
    \textbf{Technique} &
    \textbf{Acc.} & \textbf{Len.} &
    \textbf{Acc.} & \textbf{Len.} &
    \textbf{Acc.} & \textbf{Len.} &
    \textbf{Acc.} & \textbf{Len.} &
    \textbf{Acc.} & \textbf{Len.} &
    \textbf{Acc.} & \textbf{Len.} \\
    \midrule
    ConciseRL                  & 78.5 &   403 & 77.2 &  2519 & 19.9 &  4299 & 28.3 &  7670 & 29.6 &  3141 & 46.7 & 3606.4 \\
    Eff. Reasoning $\alpha{=}0.9$       & 14.6 &    15 & 18.8 &   151 & 14.6 &    18 & 27.7 &    40 & 16.5 &    19 & 18.4 &   48.6 \\
    Eff. Reasoning $\alpha{=}0.6$       & 81.3 &   655 & 77.6 &   676 & 24.3 &  1354 & 31.2 &  1252 & 34.1 &   214 & 49.7 &  727.8 \\
    Eff. Reasoning $\alpha{=}0.4$       & 60.9 &   178 & 68.0 &  2080 & 18.8 &  4126 & 31.5 &  8082 & 33.1 &  3056 & 42.5 & 3504.4 \\
    Eff. Reasoning $\alpha{=}0.2$       & 81.3 &   655 & 77.6 &  3291 & 24.3 &  7498 & 31.2 & 12078 & 34.1 &  6924 & 49.7 & 6089.2 \\
    Eff. Reasoning $\alpha{=}0.1$       & 85.0 &  1266 & 82.0 &  4226 & 20.4 &  9117 & 30.4 & 18408 & 36.1 &  7997 & 50.8 & 8202.8 \\
    Cosine Reward                       & 43.6 & 21887 & 44.4 & 22320 & 17.8 & 28788 & 28.8 & 31027 & 13.6 & 30430 & 29.6 & 26890.4 \\
    DeepScaleR                    & 80.7 & 5044  & 82.6 & 6160 & 31.5 & 10071 & 32.6 & 17742 & 34.8 & 10970 & 52.4 & 9997.4 \\
    Full Reasoning                      & 83.1 &  3656 & 78.8 &  7470 & 32.0 & 13310 & 31.0 & 22398 & 29.8 & 15219 & 50.9 & 12410.6 \\
    \bottomrule
  \end{tabular}%
  }
  \caption{Comparison of accuracy (higher is better) and token length (raw token count; lower is better) across datasets for STILL-3-1.5B-preview \cite{Slow_Thinking_with_LLMs_3_Preview, Slow_Thinking_with_LLMs_2}. For ConciseRL, we use GPT-4.1 mini \cite{openaigpt41} as the conciseness judge.}

  \label{tab:exact_results_still_exp_raw}
\end{table*}

\subsection{Kullback–Leibler Divergence}
\label{appendix:KL_divergence}

\begin{figure*}[ht]
    \includegraphics[width=\linewidth]{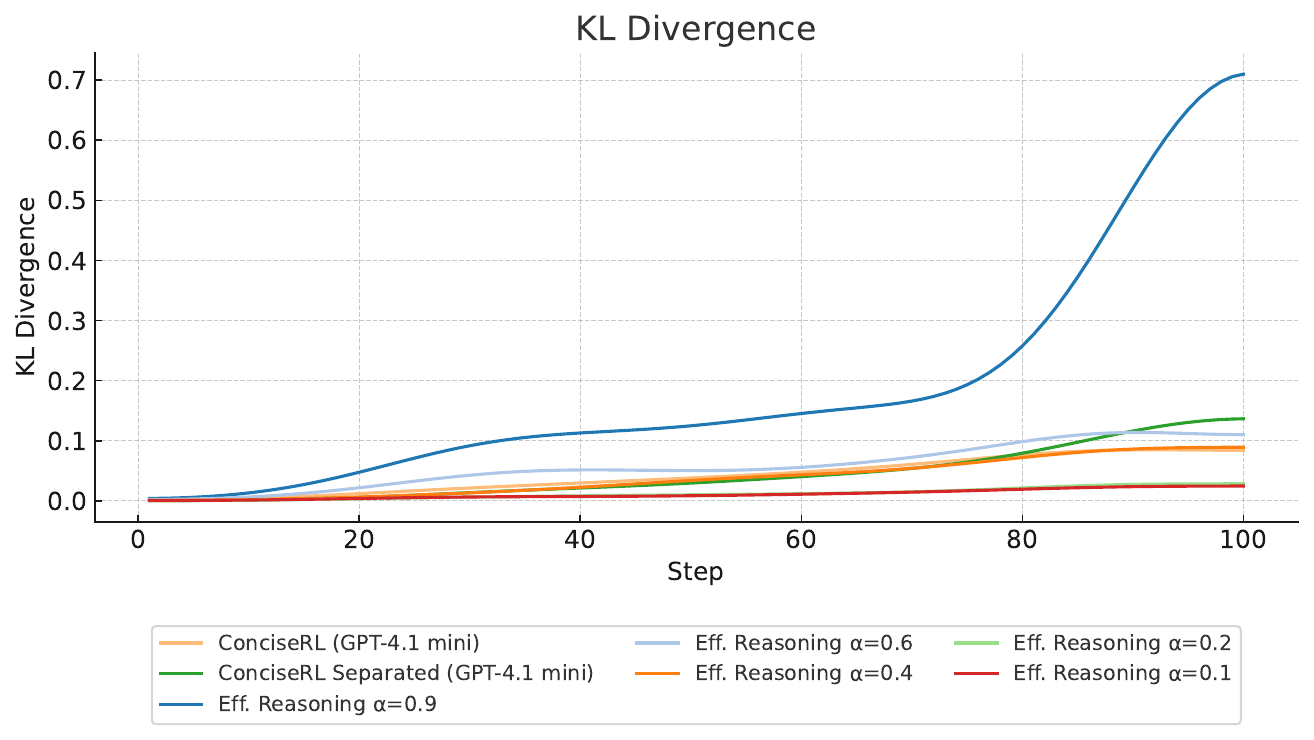}

    \caption{KL divergence between the updated policy and the initial models throughout training.}
    \label{fig:KL_divergence}
\end{figure*}

Figure~\ref{fig:KL_divergence} shows that our method (ConciseRL) induces a KL divergence profile comparable to Efficient Reasoning with $\alpha{=}0.4$, suggesting that our method modifies the policy to a similar degree while achieving significantly better efficiency–accuracy trade-offs. Also, ConciseRL results in far less KL shift than $\alpha{=}0.6$, which exhibits early instability and overshooting. Efficient Reasoning with $\alpha{=}0.9$ produces similarly short reasoning traces but incurs a much higher KL divergence, around 7$\times$ of our method, showing inefficient and unstable updates. In contrast, our ConciseRL (Separated) variant maintains low KL divergence throughout training, further reinforcing the benefits of our semantic reward and highlighting that trace length alone does not imply effective or stable policy learning.

\subsection{Future Work Discussion}
\label{appendix:future_work}

Although our method rewards semantic conciseness, L1 uses explicit length-based RL objectives to precisely control token-level output length. This difference in objectives means that the two approaches could complement each other, enabling simultaneous control of the token budget and semantic efficiency without interference. Length-based scoring approaches, however, inherently conflict with L1's exact or maximum length constraints since they introduce competing reward signals focused solely on token minimization rather than strict adherence to specified token budgets. Therefore, integrating length-based scoring directly alongside L1 is impractical, highlighting the value of our conciseness approach as a complementary strategy to optimize reasoning efficiency and performance.
Unfortunately, running such experiments requires significant compute, at least 8$\times$NVIDIA A100 80GB GPUs even for a 1.5B model, which we do not have access to.

\subsection{Conciseness Evaluation Prompt}
\label{appendix:conciseness_prompt}

\begin{figure*}[ht]
    \includegraphics[width=\linewidth]{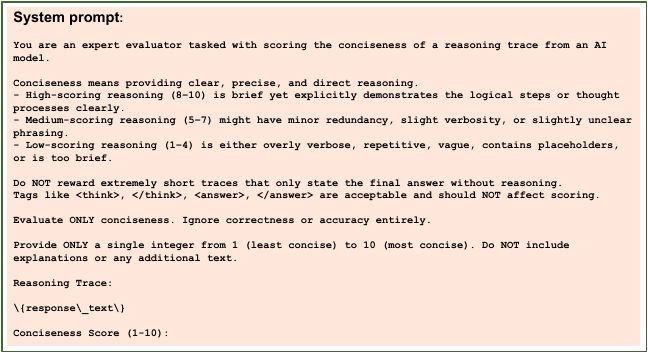}
    \caption{System propmpt used for the LLM conciseness judge.}
    \label{fig:system_prompt}
\end{figure*}

To evaluate the conciseness of reasoning traces, we use an LLM-based judge that assigns a score between 1 (overly verbose) and 10 (maximally concise), independent of correctness. The system prompt provided to the judge is designed to guide it toward assessing semantic compactness rather than surface-level brevity. The exact prompt we used can be seen in Figure~\ref{fig:system_prompt}.

\subsection{Reasoning Trace Examples}
\label{appendix:reasoning_trace_examples}

To better illustrate the differences between reward strategies, we provide representative reasoning traces from ConciseRL, ConciseRL (Separated), and Efficient Reasoning with $\alpha=0.4$. We selected $\alpha=0.4$ because it achieves accuracy comparable to ConciseRL on MATH500, making it a fair baseline for direct comparison. We focus on examples where all methods correctly solve the problem to ensure the comparison of conciseness is not confounded by correctness. For more aggressive penalty settings like $\alpha=0.6$ or $\alpha=0.9$, it was difficult to find consistent examples across all levels that resulted in correct answers, so they were excluded.

The conciseness judge rewards {\em semantic density}: a trace must enumerate every logically necessary step while avoiding rhetorical or computational detours. Because the reward is assigned by a large‐model evaluator, it is sensitive to redundancy that naive length penalties overlook, yet agnostic to correctness, which we enforce separately. This design choice underlies all the gains discussed below.

\begin{figure*}[ht]
    \includegraphics[width=\linewidth]{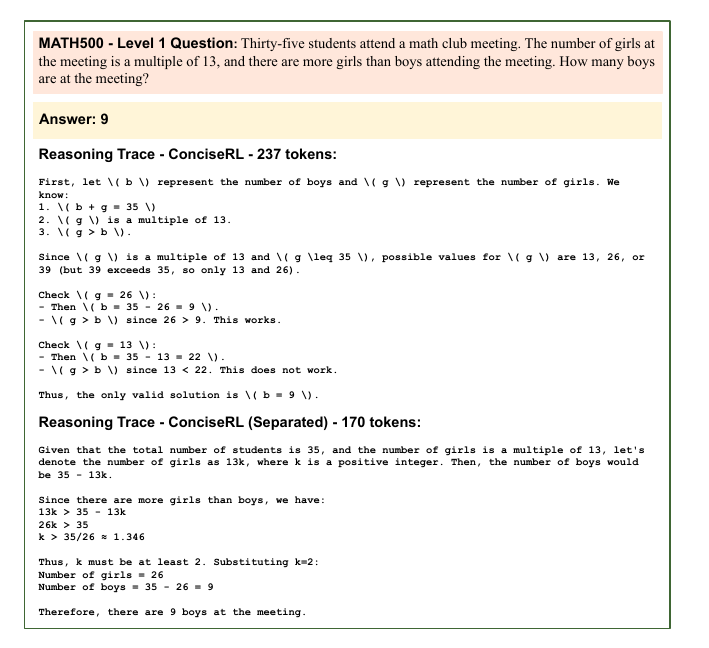}
    \caption{Example reasoning traces generated by our methods ("ConciseRL" and "ConciseRL (Separated)") on a Level 1 MATH500 \cite{hendrycks2021measuringmathematicalproblemsolving} question.}
    \label{fig:level1_ours}
\end{figure*}

\begin{figure*}[ht]
    \includegraphics[width=\linewidth]{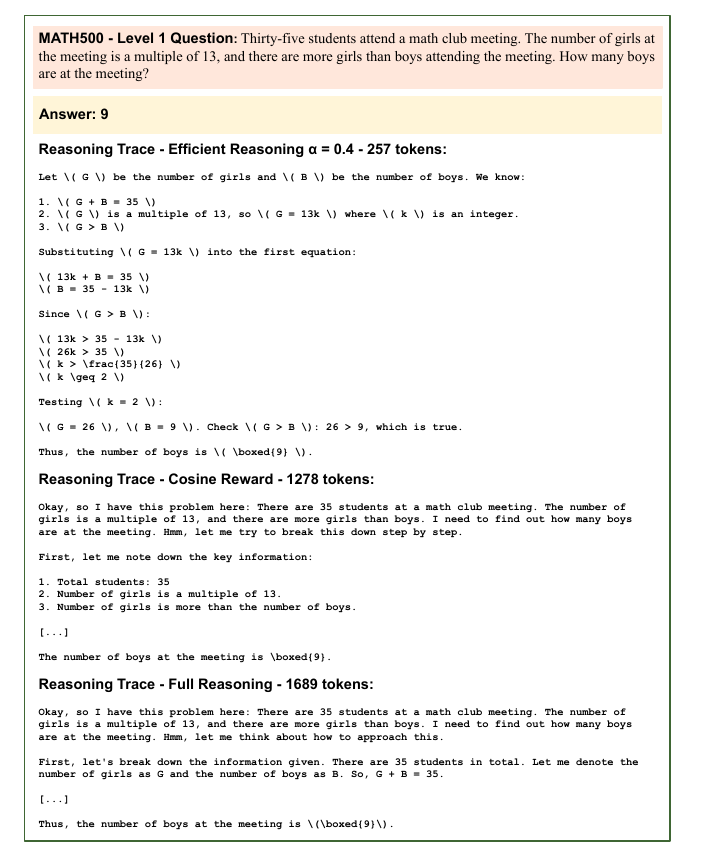}
    \caption{Example reasoning traces for a Level 1 MATH500 \cite{hendrycks2021measuringmathematicalproblemsolving} question across different methods. We show Efficient Reasoning \cite{arora2025traininglanguagemodelsreason} with $\alpha=0.4$ since it achieves the best accuracy-length trade-off for their method. The Cosine Reward \cite{yeo2025demystifyinglongchainofthoughtreasoning} and Full Reasoning \cite{deepseekai2025deepseekr1incentivizingreasoningcapability} outputs are considerably longer, so traces are partially truncated with \texttt{[...]} due to space. The full reasoning traces are available at https://github.com/RazvanDu/ConciseRL.}
    \label{fig:level1_rest}
\end{figure*}

\paragraph{Level-1 Example (\autoref{fig:level1_ours}–\autoref{fig:level1_rest}).}
Our {\sc ConciseRL} trace solves the parity-of-students puzzle in \textbf{237} tokens, with the gated variant using only \textbf{170}. Efficient-Reasoning ($\alpha$ = 0.4) needs \textbf{257}, while Cosine-Reward expands to \textbf{1278}. Most of the extra tokens in the baselines are meta-commentary or re-introductions of already stated variables; our trace states the constraints once, enumerates the only two admissible multiples of 13, and dismisses the invalid option in a single inequality check, yielding length savings without omitting any logical step.

\begin{figure*}[ht]
    \includegraphics[width=\linewidth]{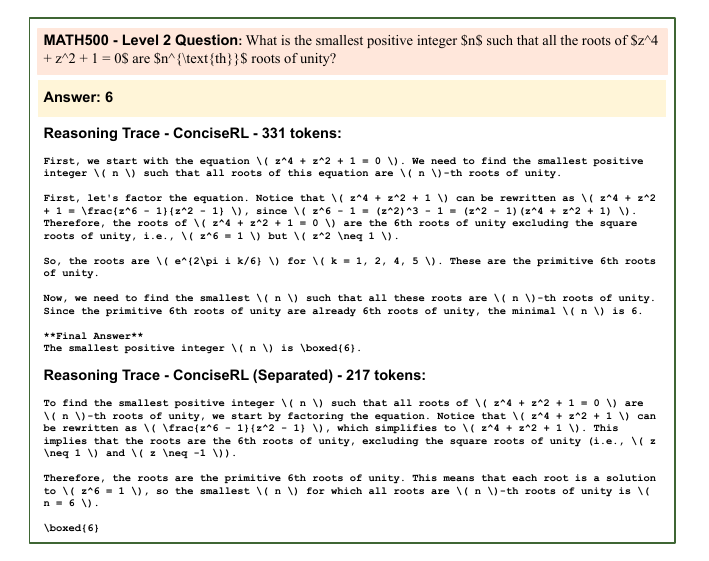}
    \caption{Example reasoning traces generated by our methods ("ConciseRL" and "ConciseRL (Separated)") on a Level 2 MATH500 \cite{hendrycks2021measuringmathematicalproblemsolving} question.}
    \label{fig:level2_ours}
\end{figure*}

\begin{figure*}[ht]
    \includegraphics[width=\linewidth]{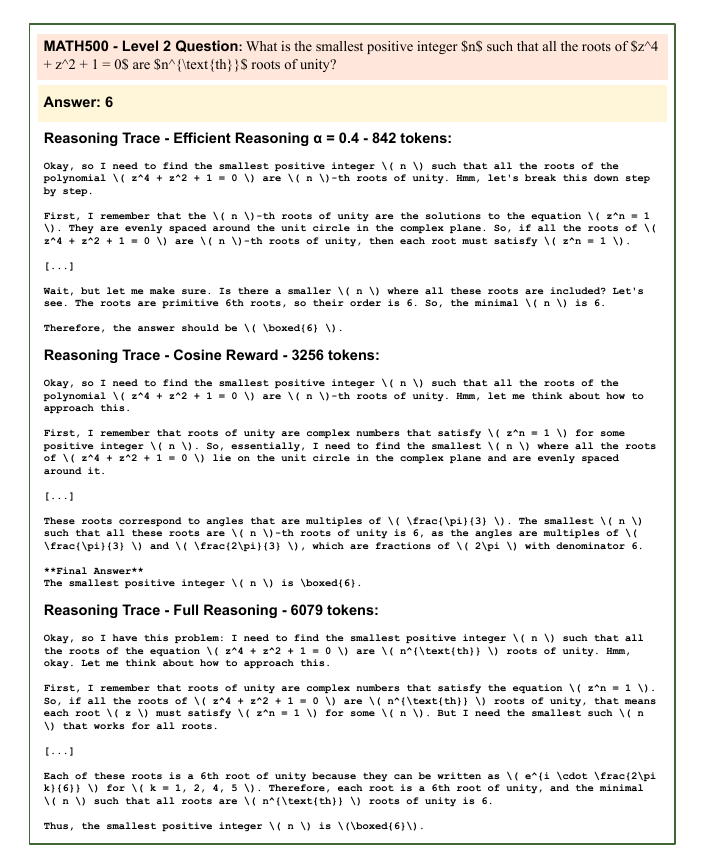}
    \caption{Example reasoning traces for a Level 2 MATH500 \cite{hendrycks2021measuringmathematicalproblemsolving} question across different methods. We show Efficient Reasoning \cite{arora2025traininglanguagemodelsreason} with $\alpha=0.4$ since it achieves the best accuracy-length trade-off for their method. The Efficient Reasoning \cite{arora2025traininglanguagemodelsreason}, Cosine Reward \cite{yeo2025demystifyinglongchainofthoughtreasoning} and Full Reasoning \cite{deepseekai2025deepseekr1incentivizingreasoningcapability} outputs are considerably longer, so traces are partially truncated with \texttt{[...]} due to space. The full reasoning traces are available at https://github.com/RazvanDu/ConciseRL.}
    \label{fig:level2_rest}
\end{figure*}

\paragraph{Level-2 Example (\autoref{fig:level2_ours}–\autoref{fig:level2_rest}).}
For the “roots of unity’’ problem, {\sc ConciseRL} finishes in \textbf{331} tokens (and \textbf{217} for the separated reward), whereas Efficient-Reasoning $\alpha$ = 0.4 requires \textbf{842}. The baselines repeatedly paraphrase the factorisation \(z^6-1=(z^2-1)(z^4+z^2+1)\), reformulate roots of unity three times, and speculate about smaller \(n\); our model invokes the factorisation exactly once, recognises that the remaining roots are primitive sixth roots, and concludes immediately. This trims the narrative overhead while keeping the algebra explicit.

\begin{figure*}[ht]
    \includegraphics[width=\linewidth]{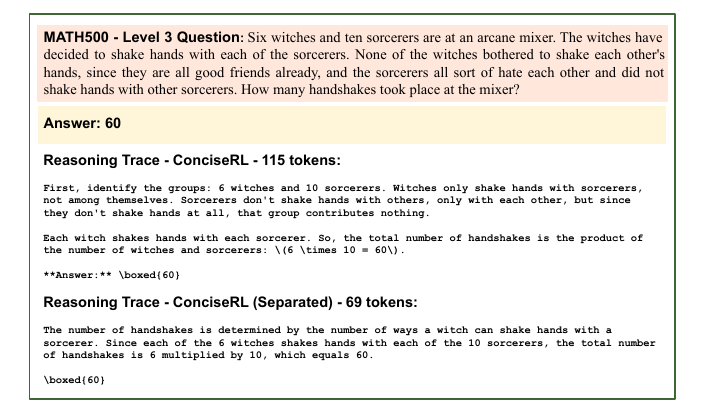}
    \caption{Example reasoning traces generated by our methods ("ConciseRL" and "ConciseRL (Separated)") on a Level 3 MATH500 \cite{hendrycks2021measuringmathematicalproblemsolving} question.}
    \label{fig:level3_ours}
\end{figure*}

\begin{figure*}[ht]
    \includegraphics[width=\linewidth]{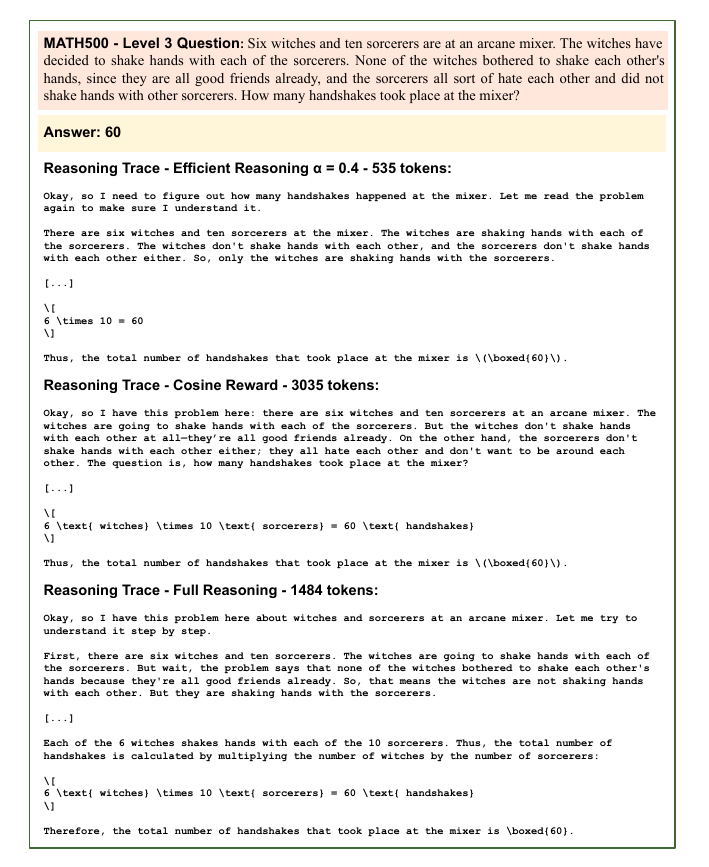}
    \caption{Example reasoning traces for a Level 3 MATH500 \cite{hendrycks2021measuringmathematicalproblemsolving} question across different methods. We show Efficient Reasoning \cite{arora2025traininglanguagemodelsreason} with $\alpha=0.4$ since it achieves the best accuracy-length trade-off for their method. The Efficient Reasoning \cite{arora2025traininglanguagemodelsreason}, Cosine Reward \cite{yeo2025demystifyinglongchainofthoughtreasoning} and Full Reasoning \cite{deepseekai2025deepseekr1incentivizingreasoningcapability} outputs are considerably longer, so traces are partially truncated with \texttt{[...]} due to space. The full reasoning traces are available at https://github.com/RazvanDu/ConciseRL.}
    \label{fig:level3_rest}
\end{figure*}

\paragraph{Level-3 example (\autoref{fig:level3_ours}–\autoref{fig:level3_rest}).}
In the handshake-counting puzzle, {\sc ConciseRL} compresses the reasoning to \textbf{115} tokens (or \textbf{69} in the separated variant) versus \textbf{535} for Efficient-Reasoning and \textbf{3035} for Cosine-Reward. The baselines re-read the prompt, rehearse elementary combinatorics, and pepper in “thinking-aloud’’ asides; our trace states the bipartite handshake structure in one declarative sentence and performs the \(6\times10\) multiplication once, making the combinatorial insight transparent.

\begin{figure*}[ht]
    \includegraphics[width=\linewidth]{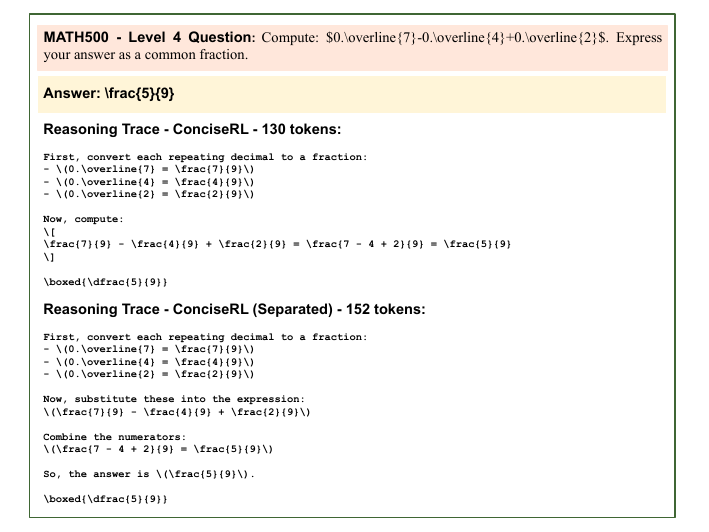}
    \caption{Example reasoning traces generated by our methods ("ConciseRL" and "ConciseRL (Separated)") on a Level 4 MATH500 \cite{hendrycks2021measuringmathematicalproblemsolving} question.}
    \label{fig:level4_ours}
\end{figure*}

\begin{figure*}[ht]
    \includegraphics[width=\linewidth]{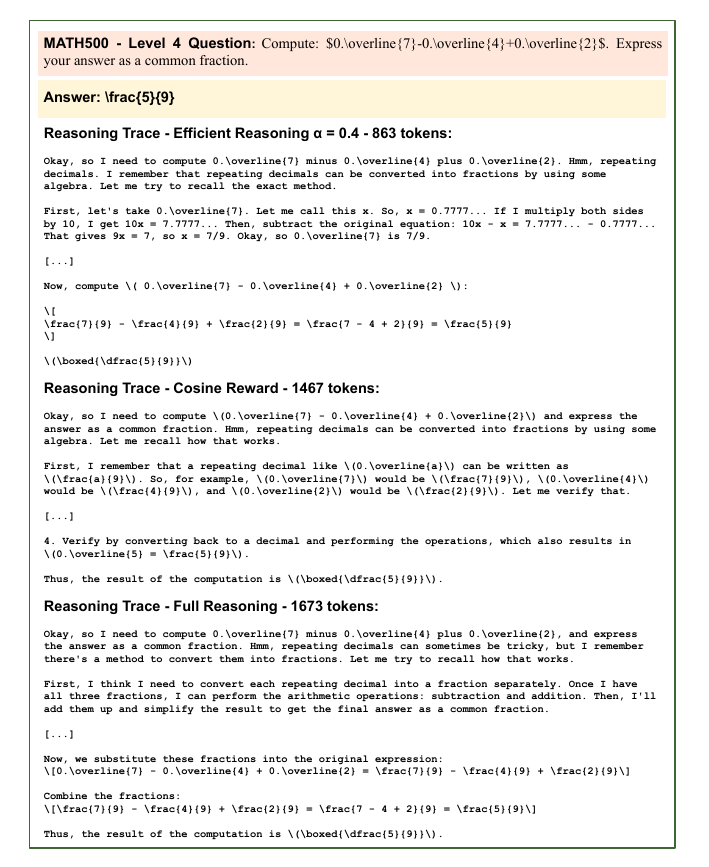}
    \caption{Example reasoning traces for a Level 4 MATH500 \cite{hendrycks2021measuringmathematicalproblemsolving} question across different methods. We show Efficient Reasoning \cite{arora2025traininglanguagemodelsreason} with $\alpha=0.4$ since it achieves the best accuracy-length trade-off for their method. The Efficient Reasoning \cite{arora2025traininglanguagemodelsreason}, Cosine Reward \cite{yeo2025demystifyinglongchainofthoughtreasoning} and Full Reasoning \cite{deepseekai2025deepseekr1incentivizingreasoningcapability} outputs are considerably longer, so traces are partially truncated with \texttt{[...]} due to space. The full reasoning traces are available at https://github.com/RazvanDu/ConciseRL.}
    \label{fig:level4_rest}
\end{figure*}

\paragraph{Level-4 Example (\autoref{fig:level4_ours}–\autoref{fig:level4_rest}).}
When converting repeating decimals to fractions, {\sc ConciseRL} requires only \textbf{130} tokens, while Efficient-Reasoning consumes \textbf{863} and Cosine-Reward \textbf{1467}. All three traces apply identical fraction conversions; our model simply avoids restating intermediate results and deletes free-form reflections such as “Hmm, repeating decimals can be tricky,’’ leading to reduction in length with no loss of clarity.

\begin{figure*}[ht]
    \includegraphics[width=\linewidth]{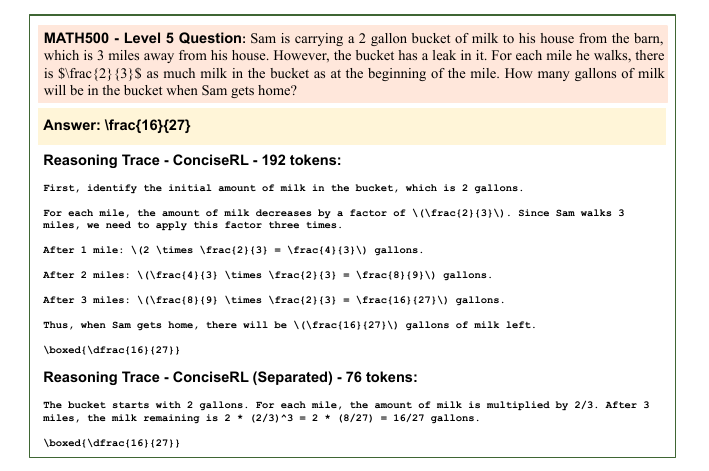}
    \caption{Example reasoning traces generated by our methods ("ConciseRL" and "ConciseRL (Separated)") on a Level 5 MATH500 \cite{hendrycks2021measuringmathematicalproblemsolving} question.}
    \label{fig:level5_ours}
\end{figure*}

\begin{figure*}[ht]
    \includegraphics[width=\linewidth]{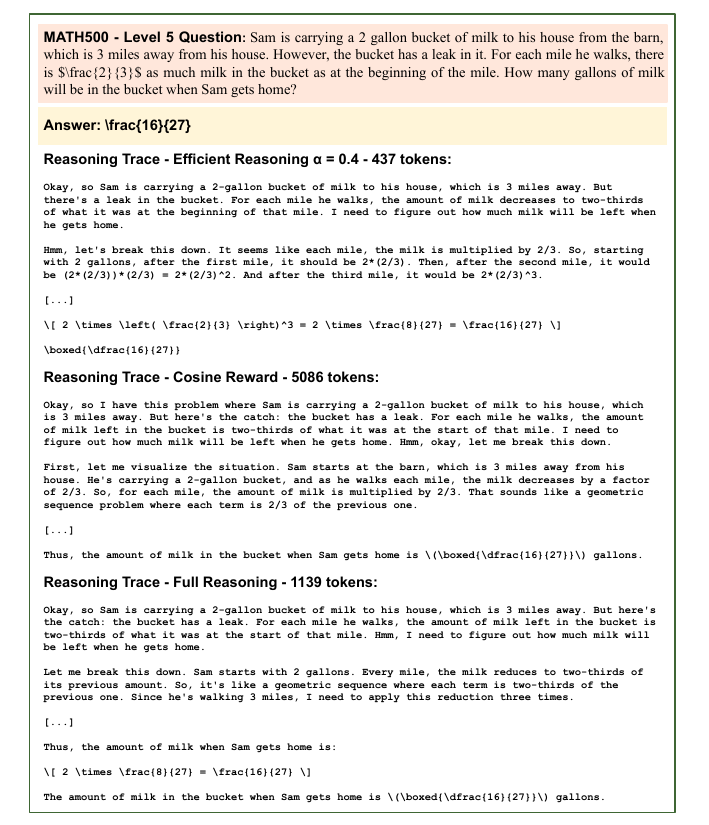}
    \caption{Example reasoning traces for a Level 5 MATH500 \cite{hendrycks2021measuringmathematicalproblemsolving} question across different methods. We show Efficient Reasoning \cite{arora2025traininglanguagemodelsreason} with $\alpha=0.4$ since it achieves the best accuracy-length trade-off for their method. The Efficient Reasoning \cite{arora2025traininglanguagemodelsreason}, Cosine Reward \cite{yeo2025demystifyinglongchainofthoughtreasoning} and Full Reasoning \cite{deepseekai2025deepseekr1incentivizingreasoningcapability} outputs are considerably longer, so traces are partially truncated with \texttt{[...]} due to space. The full reasoning traces are available at https://github.com/RazvanDu/ConciseRL.}
    \label{fig:level5_rest}
\end{figure*}

\paragraph{Level-5 Example (\autoref{fig:level5_ours}–\autoref{fig:level5_rest}).}
For the leaking-bucket problem, {\sc ConciseRL} uses \textbf{192} tokens (or \textbf{76} separated) versus \textbf{437} for Efficient-Reasoning, \textbf{5 086} for Cosine-Reward, and \textbf{1139} for full reasoning. Baselines repeatedly explain geometric decay and rewrite the same power of \(\tfrac23\); our trace articulates the multiplicative factor once and streams three intermediate states inline, preserving every quantitative step while cutting the filler.

\paragraph{Emergent Pattern across Difficulty Levels.}
Across Levels 1–5, {\sc ConciseRL} consistently removes filler phrases, nested restatements, and speculative digressions, leaving only the algebraic core. The percentage of tokens saved grows with problem difficulty, because long chains of identical transformations are where verbosity compounds. Importantly, the resulting traces are {\em more} readable: each symbol is introduced once, variables are reused consistently, and the final boxed answer is reached without back-tracking. The concision gain does not come at the expense of policy stability; Appendix \ref{appendix:KL_divergence} shows that our KL-divergence curve mirrors that of Efficient-Reasoning $\alpha$ = 0.4 while producing far shorter traces and higher average accuracy.

\paragraph{Why Our Model is Better.}
The qualitative evidence above corroborates the quantitative results from the paper: by optimising a {\em semantic} conciseness reward, {\sc ConciseRL}
(i) achieves {\bf higher accuracy} at matched or lower token budgets, because unnecessary tokens often accompany spurious logical branches that hurt correctness;
(ii) offers {\bf superior interpretability}, as every surviving token carries deductible mathematical purpose, enabling rapid human or downstream-model verification; and
(iii) incurs {\bf lower inference cost}—up to 31x fewer generated tokens on easy problems—without extra hyper-parameters or dataset-specific tuning. In short, our reward converts verbosity into signal, aligning the policy with human preferences for clear, economical argumentation.

\end{document}